# Strategic Alert Throttling for Intrusion Detection Systems

Gianni Tedesco and Uwe Aickelin
The School of Computer Science & IT
The University of Nottingham
Jubilee Campus, Wollaton Road, Nottingham
United Kingdom
{gxt,uxa}@cs.nott.ac.uk

*Abstract:* - Network intrusion detection systems are themselves becoming targets of attackers. Alert flood attacks may be used to conceal malicious activity by hiding it among a deluge of false alerts sent by the attacker. Although these types of attacks are very hard to stop completely, our aim is to present techniques that improve alert throughput and capacity to such an extent that the resources required to successfully mount the attack become prohibitive. The key idea presented is to combine a token bucket filter with a real-time correlation algorithm. The proposed algorithm throttles alert output from the IDS when an attack is detected. The attack graph used in the correlation algorithm is used to make sure that alerts crucial to forming strategies are not discarded by throttling.

*Key-Words:* - Intrusion Detection Systems, Intrusion Alert Correlation, Attack Graphs, Denial of Service Attacks, Token Bucket Filter

## 1  Introduction

As global awareness of information security issues has increased, so has the proliferation of intrusion detection technology. Network intrusion detection systems (NIDSs or simply IDSs) are quickly becoming a crucial part of the Internet security infrastructure.

Back in March 2001, there was a media furore[1] when the FBI Internet crime division issued a warning concerning the then unreleased Stick[2] program which "essentially disarms intrusion detection systems." The tool automated what we shall call the alert flood attack.

The attack works because each time an intrusion detection system raises an alert it must make some attempt to communicate the information to an operator. This communication channel can therefore become the target of a denial of service attack because, like all communication channels, it has a fixed capacity. If this channel can become overwhelmed with bogus data, an attacker can quickly achieve complete neutralization of intrusion detection capability.

There are, in fact, numerous possible types of denial of service attack against a network IDS[3], but we will focus on this particular attack type.

A great deal of research has gone in to techniques for reducing false positive alarms generally. One such technique is alert correlation. The aim of alert correlation is to analyse the alert stream and discover strategies or attack scenarios using some kind of model of possible attacker strategies[4]. One quite intuitive type of model is an attack graph[5,6,7]. The advantage of this kind of correlation is that alerts which do not (yet) conform to a threatening attack strategy are not displayed.

We propose a novel algorithm to protect NIDSs from alert-flooding attacks. The algorithm combines a throttling algorithm, namely a token bucket filter, with an existing real-time alert correlation algorithm. The aim is to reduce alerting throughput in the face of an alert flood attack, while minimising the chances of missing important alerts. The key to our approach is using the attack graph to inform the throttling algorithm so that they key alerts which make up threatening strategies are not dropped by the the sensor.

The next section of this paper will present the relevant background for the proposed techniques. The alert flood attack is defined and current approaches are examined. The real-time correlation algorithm our solution is based on is also introduced. In section 3, a modified correlation algorithithm is presented which uses throttling techniques to curb alert flood attacks. In section 4 some experimental data is presented in order to demonstrate the effectiveness of our technique. We finish by presenting a summary and some concluding remarks.

## 2 Background

The pattern matching[8] model is currently the most commonly used methodology for detecting intrusion attempts. In this model the NIDS is configured with a database of known attack patterns (also called signatures). An example of a signature is shown in Listing 1. This signature alerts on traffic generated by the well-known "BackOrifice" trojan horse program and detects any incoming packets destined to user datagram protocol (UDP) port 31337, containing a specific sequence of bytes anywhere within its payload.

```
alert udp $EXTERNAL_NET any ->
 $HOME_NET 31337 (msg:"BACKDOOR
 BackOrifice access";
 content: "|ce63 d1d2 16e7
 13cf39a5 a586|";)
```

Listing 1: A Sample Rule as used e.g. by Snort.

### 2.1 Alert Flooding

Alert flooding attacks are achieved by transmitting packets that simulate intrusion attempts and which the IDS will recognise as true attacks. Taking the example signature in Listing 1, an attacker must craft a UDP packet, set the destination port to 31337, include the sequence of bytes given in the signature and flood the target network with these packets.

The possible ramifications of this type of attack against an IDS are threefold:

1. Sensor storage becomes full, preventing further logging.

2. Sensor exceeds maximum alert throughput, causing alerts to be lost, or the sensor to cease functioning.

3. The analyst becomes deluged with false information and becomes unable to distinguish real attacks from the false ones.

Because of this, attackers may use the alert flood attack as a way to conceal genuine malicious activities.

The alert flooding technique has been automated, and hence popularised, by tools such as Stick and Snot [9] which read in signatures directly from the freely available Snort [10] IDS. Each packet sent could also have crucial fields such as source and destination address modulated by adding random data into them. This random noise makes it difficult to block the attack using a simple packet filter or firewall.

Alert floods can also be exacerbated by the poor alerting performance of IDS systems in general. A quick examination of the Snort system reveals that, in its preferred output mode (called "unified"), Snort flushes its buffers needlessly in at least two places. This causes a reduction in the effectiveness of the buffering and on UNIX like systems results in added system call overhead for every logged alert.

Performance in this area can be understandably overlooked by the IDS system designer. After all, good engineering practice tells us to optimise for the common case, and, in the world of intrusion detection, an alert is not usually the common case. In fact, on a high-speed network it should be a very rare event indeed.

Perhaps the simplest way to reduce data output while maintaining the same intrusion detection capability is to make minor modifications to the signatures to make sure that the IDS is as terse as possible. Such modifications are often used to reduce the number of false positive alerts generated. In fact generally speaking, signatures are usually a subtle compromise between allowing false negative and false positive alerts.

One way to make the IDS less verbose is to fine-tune signatures to examine only those packets destined for the relevant hosts. Let us consider BIND, DNS server software infamous for its security vulnerabilities. In this situation, the signatures may be modified to only look for BIND exploits if the destination address on the packet matches a pre-defined list of DNS servers. Of course, the operator may actually be interested to know that someone is attempting a BIND exploit on a workstation or a web server. That is to say, this approach tips the false alarm compromise towards the false negative side. Interestingly this problem also comes up when designing attack graph for correlation algorithms.

The Snort team addressed the problems of wide spread proliferation of automated alert flooding tools like Stick and Snot in their 1.8 release. Their solution was to implement a Transmission Control

Protocol (TCP) state tracking system which they called "stream4".

By keeping track of TCP connection states, stream4 is able to ignore any segments which are not part of such a conversation. In order to make the IDS raise an alert the attacker is now forced to transmit at least three segments, rather than just one. More importantly, because the three-way handshake requires two hosts to be communicating, the external attacker must find a host on the monitored network willing to participate. This might be prevented by a firewall blocking connections.

Currently most systems keep track of TCP states. This is mainly to protect against desynchronisation attacks such as those described by Ptacek and Newsham[3], but there is also the additional benefit of making sure that there is no such short cut in carrying out an alert flooding attack. Further to performing TCP state tracking, it is also possible to track any application layer state, enabling us to remove shortcuts even for protocols running over stateless transports such as UDP.

While this is a definite improvement, it cannot cover all cases: For example, some signatures must ignore state information as some exploits can exist as a single packet (i.e. statelessly); or because in other cases, they work over inherently stateless protocols. As we describe in the next section, token bucket filters combined with attack graph correlation can improve the situation.

## 2.2 Token Bucket Filter

A token bucket filter is an algorithm for controlling the rate of flow of data. Token bucket filters have traditionally been used in a number of applications where rate limiting has been needed. Some good examples are:

1. Network bandwidth management systems[11].

2. Flood protection in network chat / text conferencing systems such as Internet Relay Chat.

3. Flow control in network transport protocols [12].

4. Flood protection for programs that log externally generated events such as UNIX syslog.

A token bucket filter has two parameters, bucket size, and token rate [13].

Tokens are generated at the token rate and stored in a buffer called the "bucket" If the bucket becomes full, the extra tokens are just discarded. Each alert that arrives must have a token to pass through the filter. Any alert that does not have a token is called "over-limit" and does not pass the filter. If the alert rate is less than the token-rate then credit is allowed to accumulate in the bucket. This stored credit allows for the alert-rate to temporarily exceed the token rate (or "burst").

## 2.3 Attack Graph Correlation

Wang et al provide a unified approach to correlating, predicting and reasoning about missed alerts in [14]. The approach works in real-time and uses an in-memory data structure to perform the correlation. The correlation algorithm is robust in the face of missing alerts from the underlying IDS.

An in-memory data structure called a "queue graph" (QG) is introduced. In order to avoid keeping unnessecary alerts in memory, only the latest alert for a given exploit vertex is stored in this structure. That is to say that the correlation between such matching alerts is left as implicit. This allows the algorithm to be run in real-time without necessetating the usual sliding correlation window approach which would allow an attacker to use an alert flood attack to introduce false negative correlations.

In this system, attack graphs are defined as directed acyclic graphs (DAGs) having two distinct types of vertices, security conditions and exploits (see Figure 1). Exploit vertices are (vuln,src,dst) tuples. The src and dst fields are used to tie the exploit to specific combinations of vulnerable and attacking hosts, wildcards may be used. These vertices may represent one or more possible alert types. A function "f" is introduced which maps alerts to an exploit vertices in the attack graph.

Security conditions vertices refer to prerequisites and consequences of exploits. Thus edges connecting a condition to an exploit are prerequisite relations and those connecting an exploit to a condition are consequence relations.

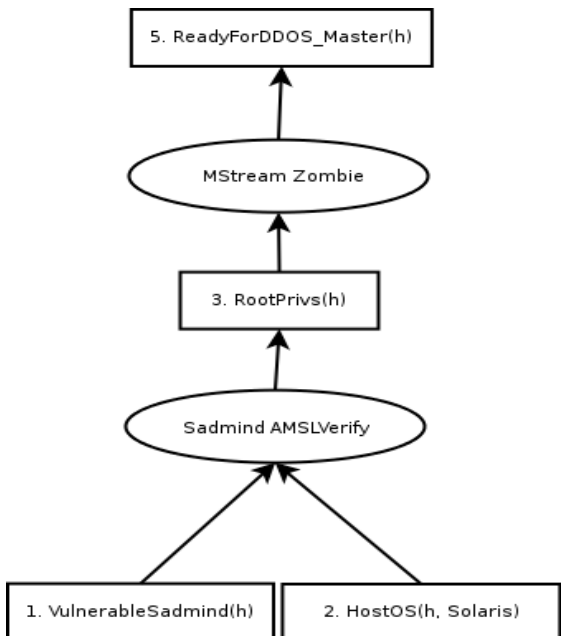

Figure 1: A Sample Attack Graph

Attack graphs are generated automatically with TVA, the topological vulnerability assessment tool[15] which links together the output of Nessus, IDS rules and a vulnerability database. In order to do this a function which maps alerts to exploits is introduced. In this way the correlation algorithm is vulnerability-centric. That is to say it will not correlate exploits against machines which are not defined as being vulnerable to them. These graphs are distinct from those used by Ning et al in that they contain not just the causal relationships between attacks but also a database of vulnerable hosts on the network.

An IDS (in this case Snort) is set up to send its alerts directly to the correlation component. The way the attack graph is used by the correlation component is to treat each exploit vertex in the graph as a queue. Alerts are placed in their requisite queue and a breadth first search is performed in the graph to find previous exploits which would correlate with the current one. If a queue is found and is non-empty then a correlation is generated. If a queue is empty, the algorithm can either stop or hypothesise a missing attack and carry on.

If the edges in the graph are directed forwards in time, rather than backwards, predictions can be generated in much the same way as correlations.

The QG structure is actually an enhanced version of the attack graph. A tree is created for each exploit vertex in the graph. In these trees, the correlation and prediction edges are all precalculated. This effectively means that correlation and prediction can be done in linear time by searching in a tree rather than quadratic time by performing breadth first search in the attack graph and this is what makes the algorithm suitable for real-time application.

The output of the algorithm is a correlation graph which can contain a mix of real and hypothesised alerts and security conditions. Readers are urged to consult the original paper for the full details[14].

## 3 Strategic Data Reduction

We have described the alert flood attack in the previous sections as fundamentally a resource exhaustion attack. In this section we will outline an approach to reduce exposure to the attack by combining alert throttling with attack graph correlation.

Consider the case of a human IDS operator as a resource that cannot cope with having to examine many thousands of bogus alerts at the rate at which a sustained attack can produce them.

There are two approaches to solving this type of problem: one is to increase the amount of resources at your disposal, the other is to reduce the amount of resources required. While it is conceivable that one could scale the sensor hardware to be fully able to cope with alert floods at a given rate for a given length of time it seems rather more complex to scale the human operator.

Taking the approach of minimising the resources required, alert data could be reduced by throttling the alert stream to a fixed rate. This could be achieved by applying a token bucket filter either per signature, per attack type, globally, or even in to complex hierarchies as in HTB3[15]. The burstiness feature of the TBF algorithm means that alerts are only discarded under sustained high rate of alerts. However such approaches run the risk of dropping important alerts which can even assist an attacker in concealing their malicious activities.

The key to our approach is to allow the correlation algorithm to interpose between the signature matching, and output components of the IDS. By doing this, a token bucket filter can be placed at each queue in the QG structure and overlimit alerts can be discarded.

In order that the user may be informed of dropped alerts we can use a kind of "run length encoding"

(RLE) to represent a string of alerts. RLE is a simple compression technique which replaces recurring sequences of symbols (called runs) with a single symbol and a run count *N*. To decompress, one simply copies the symbol into the output stream *N* times. This is an approach familiar to UNIX users who have ever tried to flood the syslog program and seen its "last message repeated *N* times" warning.

To implement RLE compression in our case, we first assume that all alerts going through the same token bucket filter are identical. Then all that is required is to add a counter to the queues in the QG data structure and increment that counter for all over-limit alerts. When there is enough credit in the token bucket to permit new alerts, we dequeue the the alert and the counter, allowing them to add a node in the output graph and to be logged to permanent storage. This allows for some minimal reconstruction of lost packets by just using the information in the attack graph.

Two questions then arise. Firstly what to do with alerts not mapping to vertices in the queue graph; and secondly what parameters to use for the token bucket filters.

For those alerts which do not map in to exploit nodes, we cannot be sure that we are missing alerts vital to some strategy. Since the QG algorithm assumes a complete attack graph anyway we could discard all such alerts. A more prodent approach is taken in our case, and that is to apply a token bucket filter to such alerts on a per-signature basis.

As for the parameters of the TBFs, for those alerts which map to vertices in the attack graph, we could drop all implicitly correlating alerts and keep the same strategies. However it is seen as a benefit to keep alerts where possible, here we envisage that token rates of greater than one or two alerts per second need not be used. For other alerts however, there is, of course, a trade-off between data fidelity and efficiency.

In the next section, we will show that this technique scales up such that it effectively nullifies the computational effect of an alert flood attack.

## 4 Empirical Data

We can perform a simple test with the Firestorm[16] system running off-line against a tcpdump[17] capture file containing an alert flood attack captured by Shmoo Group at a defcon CTF event[18]. The attack consists of a repeated ICMP flood at a rate of around 7,343 packets per second.

We perform 2 tests and in both, we have a full signature database loaded containing around 1,600 signatures, with the network data read directly from the hard disk. The test machine was a 3.2GHz Pentium-IV running Linux 2.6 with 1GB of RAM. The results shown are an average of three iterations for both runs to factor out any random fluctuations such as may be caused by disk seek latency.

The first run (#1) is a control run using firestorm + QG algorithm. The second run (#2) is identical except for the addition of token bucket filtering. Two sets of filters are used:

1. The set of filters for each exploit vertex in the attack graph.
2. The set of filters for each rule which does not map to a vertex in the attack graph.

Each of these filters is set to 2 alerts per second and a burst of 20 alerts. These parameters are rather arbitrary but are probably best set based on the operators experience of the baseline alert rate for the network.

| # | Data Size (KB) | Alerts | CPU Time | Run Time |
|---|---|---|---|---|
| 1 | 475,229 | 300,741 | 13.131 | 18.476 |
| 2 | 1,092 | 696 | 12.153 | 12.817 |

Table 1: Experimental Results.

As we can see in Table 1, the amount of data logged was reduced by several orders of magnitude and the run time decreased disproportionately to the CPU time. While the run time was reduced by around 30%, the CPU time only reduced by around 10%. This indicates that the Firestorm process is not wasting as much time waiting for I/O completion when the token bucket filter is enabled,

The number of alerts output is reduced by orders of magnitude. In the experiment the communication channel between the IDS and the operator is simply an on-disk alert spool so the available bandwitdth is high. In a real world deployment, on the other hand, it is likely that alerts would be transmitted across a network adding further latency and bandwidth constraints. In these deployments we expect even greater gains in performance.

From these results it is shown that we can effectively boost performance and therefore sensor capacity, allowing the IDS to carry on working during an alert flood rather than becoming overwhelmed and possibly exhausting the storage on the sensor. Even if the attack contained twice as many packets in the same space of time, it would not double the amount of data logged as the token rate is fixed.

## 5  Summary and Conclusions

Alert flooding is a problem that will probably always exist with intrusion detection systems and one that cannot be eliminated entirely. However, we have shown that it is possible to drastically reduce the effects by recognising an attack and throttling excess alerts.

We have further shown that real-time alert correlation algorithms can be used to provide a useful context for throtting alerts such that key attacks are not missed, such an approach solves problems with either technique used in isolation.

Without the correlation system interceding between the signature matching and alerting components of the IDS it is not possible for it to decide if alerts may be logged or not and without having strategic information available to the throttling algorithm, it could drop crucial alerts.

Further investigation is required in to producing optimal token bucket filter configurations and how best to handle those alerts which do not map on to any exploit vertices in the attack graph.